\documentclass{article}
\usepackage{ijcai18}

\usepackage{times}
\usepackage{xcolor}
\usepackage{soul}
\usepackage[utf8]{inputenc}
\usepackage[small]{caption}

\usepackage{microtype}
\usepackage{graphicx}
\usepackage{subfigure}
\usepackage{amsmath}
\usepackage{amssymb}

\title{Estimate and Replace:\\A Novel Approach to Integrating\\Deep Neural Networks with Existing Applications}
\author{
Guy Hadash$^{*1}$,
Einat Kermany$^{*1}$, 
Boaz Carmeli$^{*1}$, 
Ofer Lavi$^1$,
George Kour$^1$,
Alon Jacovi$^{1,2}$
\\ 
$^*$ Equal contribution\\
$^1$IBM Haifa Research Lab, Haifa, Israel, 
$^2$Bar-Ilan University, Ramat-Gan, Israel\\
{\tt \{guyh, einatke, boazc, oferl, alon.jacovi\}@il.ibm.com}\\
\tt gkour@ibm.com\\
}

\begin{document}
\maketitle
\begin{abstract}
Existing applications include a huge amount of knowledge that is out of reach for deep neural networks. This paper presents a novel approach for integrating calls to existing applications into deep learning architectures. Using this approach, we estimate each application's functionality with an estimator, which is implemented as a deep neural network (DNN). The estimator is then embedded into a base network that we direct into complying with the application's interface during an end-to-end optimization process. At inference time, we replace each estimator with its existing application counterpart and let the base network solve the task by interacting with the existing application. Using this \textit{‘Estimate and Replace’} method, we were able to train a DNN end-to-end with less data and outperformed a matching DNN that did not interact with the external application.
\end{abstract}

\section{Introduction}
\label{introduction}

End-to-end learning with deep neural networks (DNN) has taken the stage in the past few years, achieving state-of-the-art performance in multiple domains including computer vision \cite{szegedy2017inception}, text analysis \cite{NIPS2014_5346,P15-1001} and speech recognition \cite{xiong2016achieving}.  
End-to-end learning maps input to output based solely on data. 
However, in many tasks, enabling a DNN to interact with an existing program, knowledge, or application through its Application Programming Interface (API) can lead to a superior solution. An API defines a functionality. Accessing an API is accomplished via its input and output parameters. For example, consider the simple question: "Is 7.2 greater than 4.5?". We could answer this question by letting a DNN handle the natural language part, and accessing a simple $greater\_than(7.2, 4.5)$ API to solve the logical part. We refer to this as a \textit{hybrid solution}. This type of solution can outperform pure DNN models by using analytic techniques and closed algorithms already implemented in external applications. Such external applications do not require learning, thereby using them should reduce the total amount of training data needed.

Training a DNN to interact with an external application through its API poses an inherent difficulty: most DNN training procedures rely on different variants of gradient back-propagation. Therefore, it is only possible to train them end-to-end if all solution parts are implemented as neural networks, or at least can return a valid gradient. Consequently, all parts of the overall solution must be differentiable.

One of the approaches that enable DNNs to make use of external application functionality is that of the Neural Programmer Interface  \cite{neelakantan2016learning}. The authors implement differentiable functions that approximate the desired functionality, and implant them in the network, thus bypassing the call to the external application. Another option \cite{Liang:2016:LES:2991470.2866568} is a two-step approach that includes program induction or program generation. First, the system translates the task's input into a program i.e., a sequence of API calls. Second, it executes the program on an external engine.
Reinforcement Learning (RL) is another possible approach to learn a policy - what actions should be taken and when. When learning a policy, RL lets the environment handle the external application functionality and enables the DNN to learn the policy by splitting the learning process into a sequence of discrete decisions. Such splitting prevents the RL process from using gradient guidance.

Our approach, which we refer to as \textit{Estimate and Replace}, integrates existing non-differentiable applications into DNNs. We use an estimator subnetwork, which we call \textit{EstiLayer}, to estimate each of the non-differentiable applications. We then train the whole network, which we call \textit{EstiNet}, to comply with the EstiLayer interface during an end-to-end optimization process. At inference time, we replace each EstiLayer with its external application counterpart and let the EstiNet model solve the task by accessing external application interfaces as needed.

\textit{Estimate and Replace} streamlines the integration process. It relies on existing training data and replacing an EstiLayer with its external application counterpart rather than keeping it within the DNN as is, leads to better results at inference time.
The key to this approach is an innovative multi-task training process. It forces parts of the network, the EstiLayers, to conform to a predefined functionality, while training other parts of the network to interface with the EstiLayers correctly. Training an internal layer within a DNN to execute a predefined function poses several challenges. These are described in more detail in Section \ref{learning_to_interact}.

Our work advances existing research with the following main contributions: 1) An entirely new approach to combine DNNs with existing applications. 2) A new training method that supports end-to-end optimization while constraining part of the network to conform to a predefined functionality. 3) We show that using  this approach we can train a DNN end-to-end using less data compared to a similar architecture without component estimation and replacement.


\section{RELATED WORK}
\textbf{End-to-End Learning:} Task-specific architectures for end-to-end deep learning require large datasets and work very well when such data is available, as in the case of neural machine translation \cite{bahdanau2014neural}. General purpose end-to-end architectures, suitable for multiple tasks, include the Neural Turing Machine \cite{graves2014neural} and its successor, the Differential Neural Computer \cite{Graves2016} (DNC). There is no external application integration in these architectures.
Other architectures, such as the Neural Programmer architecture \cite{neelakantan2016learning} allow end-to-end training while constraining parts of the network to execute predefined operations by re-implementing specific operations as static differentiable components. This approach has two drawbacks. It requires re-implementation of the API in a differentiable way, which may be difficult, and it lacks the accuracy and possible scalability advantages of an external API.

\textbf{Program Induction and Program Generation:} Program induction is a different approach to interaction with external APIs. The goal is to construct a program comprising a series of operations based on the input, and then execute the program to get the results. When the input is a natural language query, as in our focus, it is possible to use semantic parsing to transform the query into a logical form that describes the program \cite{Liang:2016:LES:2991470.2866568}. Early works required natural language query-program pairs to learn the mapping. Recent works, (e.g., \cite{pasupat2015compositional}) require only query-answer pairs for training.

Other approaches include neural network-based program induction, \cite{andreas2016learning}, translation of a query into a program using sequence-to-sequence deep learning methods \cite{lin2017program}, and learning the program from execution traces \cite{reed2015neural,cai2017making}.
One major difficulty of neural methods is the need to make a discrete selection of each step in the program. Some works, (e.g., \cite{andreas2016learning}) overcome this difficulty by substituting the real gradient with an estimate of the gradient using the REINFORCE algorithm \cite{williams1992simple}.


\textbf{Reinforcement Learning:} Learning to execute the right operation can be viewed as a reinforcement learning problem. For a given input, the agent has to select the optimal action from a set of available actions. The action selection repeats following feedback based on the previous action selection. Earlier works that took this approach include \cite{Branavan:2009:RLM:1687878.1687892}, and later \cite{artzi2013weakly}. Recently, \cite{DBLP:journals/corr/ZarembaS15} proposed a reinforcement extension to Neural Turing Machines \cite{graves2014neural}. In \cite{tamar2016value}, the authors pose a value iteration based solution for reinforcement learning tasks as an end-to-end learning task with a Value Iteration Network (VIN). VIN are shown to learn how to plan a sequence of actions for a given task.

\section{ESTIMATE AND REPLACE}
\label{estinet_approach}
\begin{figure}[ht]
\vskip 0.2in
\begin{center}
\centerline{\includegraphics[width=80mm]{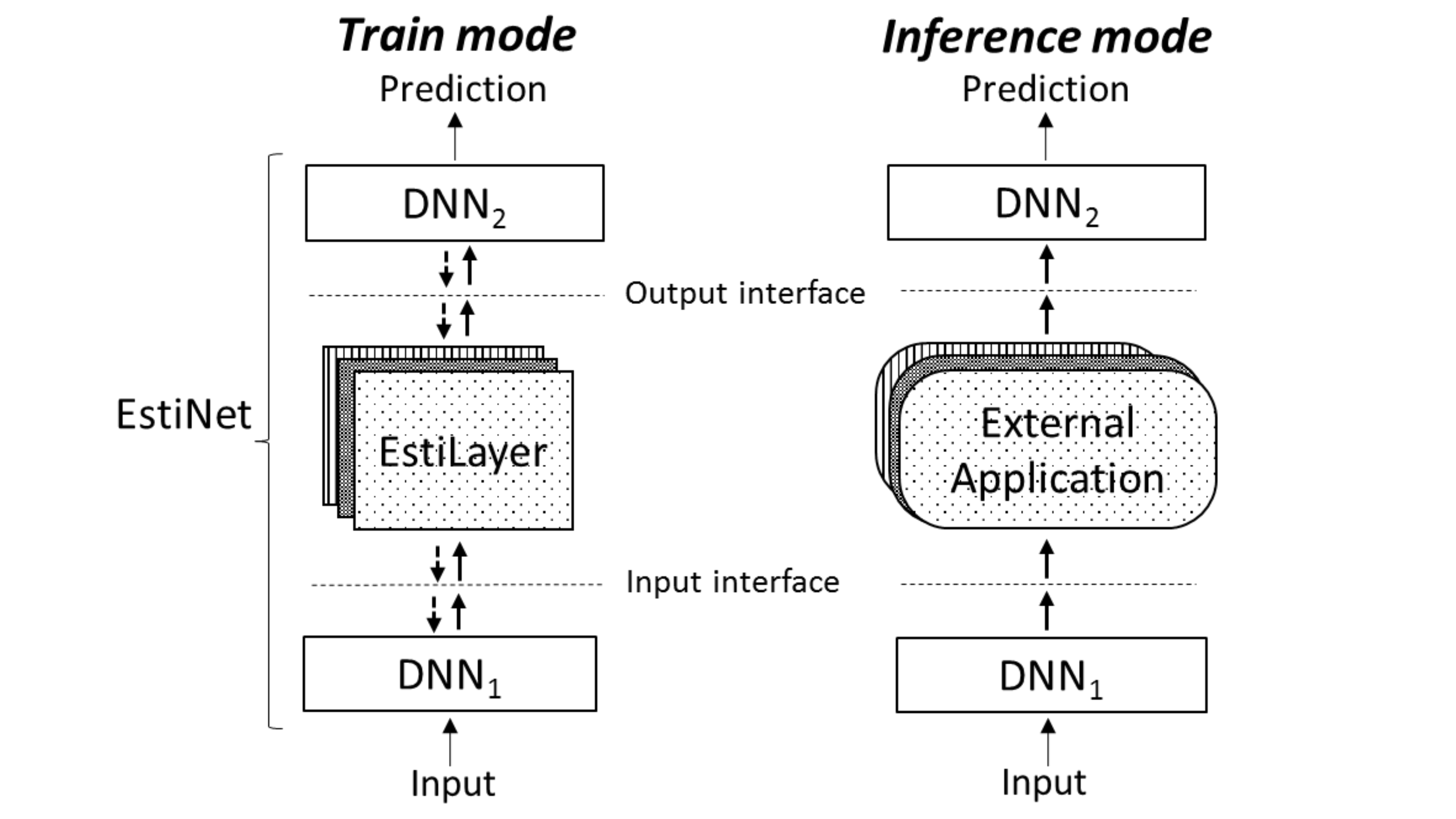}}
\caption{The \textit{Estimate and Replace} approach: Train - Use an EstiLayer to estimate an external functionality and train an EstiNet DNN model end-to-end. Inference - Replace the each EstiLayer with an external application to achieve optimal results. A rectangle represents a neural network and a rectangle with rounded corners represent an external application. Solid arrows are for the forward pass and dashed arrows are for the back propagation flow.}\label{estinet-approach}
\end{center}
\vskip -0.2in
\end{figure}

In this section we present \textit{Estimate and Replace}. Our approach enables a DNN to interact with external, non-differentiable applications, as illustrated in Figure~\ref{estinet-approach}. At the heart of this approach lie estimator subnetworks, or \textit{EstiLayers}, which we use to estimate each external application. The EstiLayers encourage the DNN model, which we refer to as an \textit{EstiNet}, to comply to the predefined EstiLayers' interface during an end-to-end optimization process. At inference time, we replace each EstiLayer with its external application counterpart, and let the EstiNet access it as needed. By replacing EstiLayers with external applications that share the same interface, we can use the strengths and advantages of each of the computational components, namely the DNN and the external application, in an accurate and efficient way.

\subsection{Interface Learning Challenges}
\label{learning_to_interact}
The Estimate and Replace approach implies a DNN ability to learn an application interface during training and use it at inference time. This approach raises three significant training challenges:
\begin{enumerate}
\item\textbf{Selecting the right interface:} To successfully complete a task, we have to train the DNN to access the right API at the right time. Selecting from a collection of APIs is a discrete operation and thus, non-differentiable. This inherent discrete API selection poses an immediate difficulty on the end-to-end training process.
\item\textbf{Constraining the interface parameters:} To replace an EstiLayer with its application counterpart at inference time, we have to constrain its interface to comply with the API input and output parameters. Constraining an EstiLayer to an input and output definition as part of an end-to-end training process poses another challenge. 
\item\textbf{Executing a sequence of API calls:} Most tasks require the execution of a sequence of API calls for successful completion. Executing a call sequence poses an additional burden on the training process. The network now needs to consume the output of an API call and adjust it to conform with the input of the successive call. Addressing this additional network orchestration difficulty is beyond the scope of this work. We plan to confront this challenge in future work.  
\end{enumerate}

\subsection{The EstiNet Model}
To address the abovementioned challenges, EstiNet defines an abstraction with three conceptual subnetworks: 1) Input representation 2) Selectors and 3) EstiLayers. These three subnetworks provide the main building blocks required to solve a specific task. Next, we describe the main functionality of each of these three subnetworks.

\subsubsection{Input Representation} EstiNet uses an input representation subnetwork \(f_{in}(x)\) to represent the task input. The task data derives the exact architecture of this subnetwork. The input representation architecture allows the selectors to extract the needed information from the input.

\subsubsection{Selectors} EstiNet uses selector subnetworks \(f_{sel}()\) for two complementary tasks: (1) To figure out the proper API call and (2) To extract the proper API arguments from the input. The selectors assign a probability distribution over a set of classes, and use Gumbel Softmax \cite{jang2017categorical} to facilitate discrete selections. The task data and the available APIs define the exact number of selectors and their specific output.

\subsubsection{EstiLayers} EstiNet uses EstiLayer subnetworks \(f_{est}()\) to estimate an API and its functionality. The main role of the EstiLayer is to estimate an external, non-differentiable application with a differentiable subnetwork, allowing us to train the network end-to-end with stochastic gradient descent (SGD). Forcing an EstiLayer to estimate the application's functionality and comply with its API allows us to replace it with the actual application at inference time, while keeping the overall functionality intact. 
In this work, we implement all EstiLayers with a general purpose DNN architecture \cite{vaswani2017attention}. EstiLayers are not limited to non-differential applications and are also valuable for estimating complex differentiable APIs. Using EstiLayers eliminates the need to directly implement a complex API, by allowing the network to learn the functionality from the data. We can then use the external application at inference time to achieve better performance. 

\subsection{Auxiliary Functionality}
Enabling EstiNet to interact with an external application requires additional functionality, which we describe below.
\subsubsection{Number Representation} 
Numerical APIs (e.g., \textit{greater-than}) use numbers as their arguments. Forcing an EstiLayer and a numerical API to have the same interface necessitates a compatible number representation. To this end, EstiNet must represent a number in a format that can be translated to and from a concrete number. Moreover, many tasks require representations for both numbers and words. It may be desirable to embed both representations into the same vector space. Previous works either represent numbers as is \cite{ling2017program} or replace them with a predefined token. In this work, we use a number-embedding approach that allows EstiNet to handle numbers and words interchangeably.

\subsubsection{Adaptation Function} EstiNet uses adaptation functions to adapt selectors output to the required API input, and to adapt API output to the required input of higher network layers at inference time. The adaptation function may be non-differentiable; thus, it cannot be used during back-propagation. Nevertheless, EstiNet uses these functions for label generation during training while training (see Section \ref{online_training} for more details).

\subsection{Training Procedures}
\label{training}
The procedure of training a DNN end-to-end, which we refer to as \textit{plain training}, is carried out by providing task labels as supervision. This supervision allows the model to learn the input-to-output mapping end-to-end, but does not provide any guidance for learning an application interface.   
We use plain training as our baseline and examine two enhanced training procedures to support the Estimate and Replace approach: \textit{offline} and \textit{online} training. Next, we describe these two procedures in detail.

\subsubsection{Offline Training} 
\textit{Offline training} is a two-step procedure. First, we train an EstiLayer to estimate an application’s functionality. We create training data by generating application input, and then record its output.
Second, we load the trained EstiLayer into the EstiNet model and train the EstiNet end-to-end while keeping the EstiLayer parameters fixed.
A procedure we refer to as \textit{offline-trainable} changes the EstiLayer functionality by allowing the optimizer to update the EstiLayer parameters during the end-to-end training process. This changes the learned interface with the selectors subnetwork. Consequently, we expect performance at inference time to decline. We report the results of these two training procedures in Section \ref{training_procedure_results}. 
\subsubsection{Online Training} 
\label{online_training}
\begin{figure}[ht]
\vskip 0.2in
\begin{center} 
\centerline{\includegraphics[width=70mm]{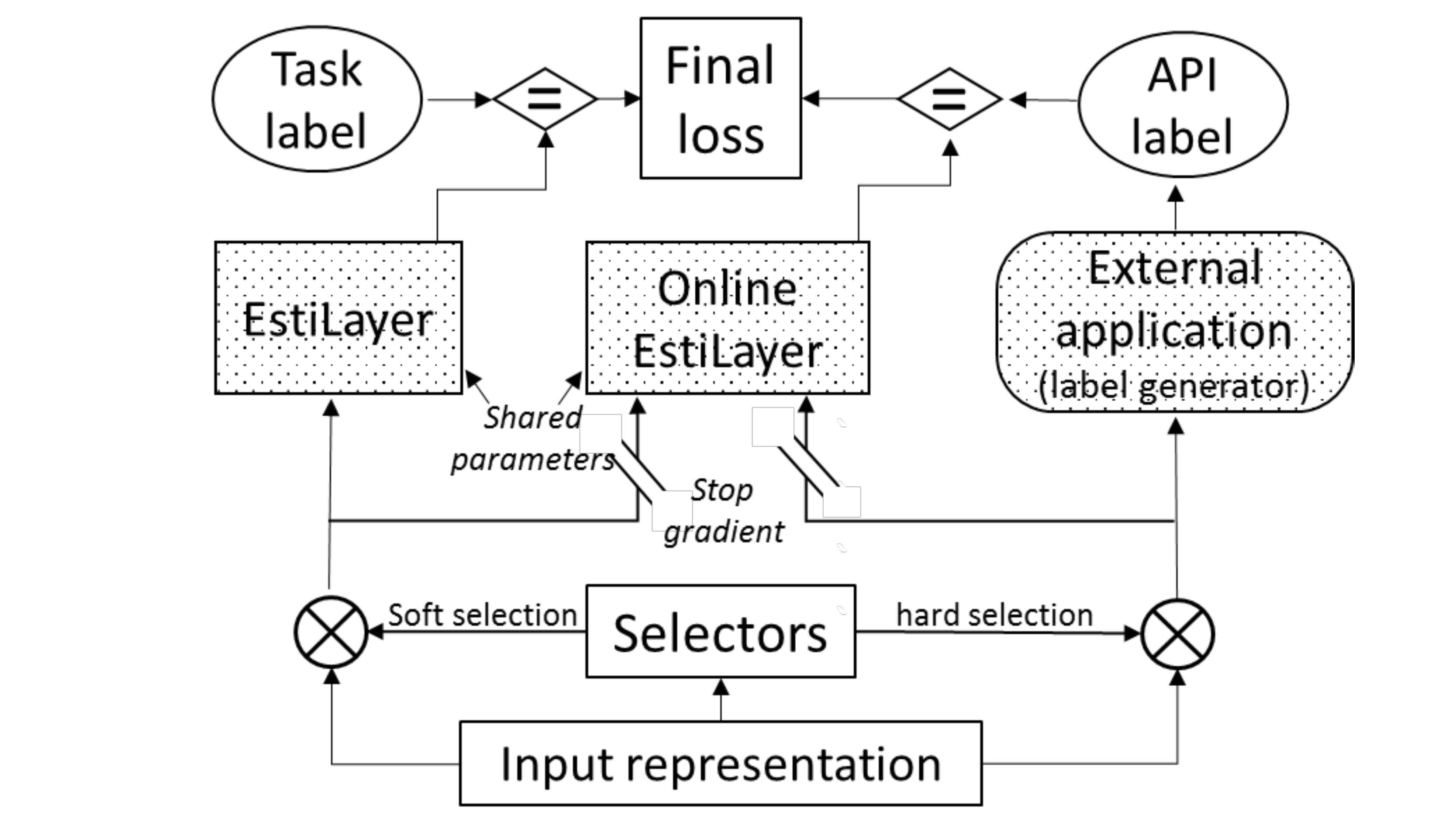}}
\caption{Label generation and online-training. We jointly optimize the model parameters with respect to the task label (left side) and the EstiLayer parameters with respect to the application label (middle and right side). We stop the gradients flow while optimizing the EstiLayer parameters with respect to the application label, and thus enforce it to retain application functionality. Note that 1) the EstiLayer and the online EstiLayer share their parameters and 2) the API label is generated online from the input to the EstiLayer and not from the task input.}
\label{online-training}
\end{center}
\vskip -0.2in
\end{figure}

\textit{Online training} is a multi-task training procedure that jointly trains the EstiLayers and the whole EstiNet model. It calculates an additional loss value, based on a label that it generates online from the external application. Let $x$ be some model input sample. Let $z_{soft}=f_{soft\_sel}(f_{in}(x))$ and $z_{hard}=f_{hard\_sel}(f_{in}(x))$ be the selectors' input to the EstiLayer interface while in soft and hard selection modes, respectively. We then define the EstiNet model prediction to be $f_{model}(x)$, the soft online prediction $f_{online\_soft}$ to be $f_{est}(z_{soft})$, and the hard online prediction $f_{online\_hard}$ to be $f_{est}(z_{hard})$. Furthermore, we define $y$ to be the task label for input sample $x$ and $y_{api}=f_{api}(f_{adp}(z_{hard}))$ to be the API label for the API input sample $z_{hard}$. We can now define the total loss $L$ to be: $$L = L_{model} + \lambda_1 L_{online\_soft} + \lambda_2 L_{online\_hard}$$ 
where $\lambda_1$ and $\lambda_2$ are hyper parameters of the model and $L_{model}=loss(f_{model}, y)$,\\ $L_{online\_soft}=loss(f_{online\_soft}, y_{api})$,\\  and $L_{online\_hard}=loss(f_{online\_hard}, y_{api})$.\\ 
Figure~\ref{online-training} presents a schematic diagram of the online training procedure. A variant of the online training procedure is \textit{online-pretraining}; here, we start by training the EstiLayers as in the offline training and then use it during the online training. This procedure yielded the best EstiNet performance, and as such, is our recommended training procedure.

\subsection{Performance Assessment}
We assessed EstiNet in two different modes: \textit{test modes} and \textit{inference mode}. In test mode we used a dedicated test set applied to the learned network. In inference mode we replaced each EstiLayer with its external application counterpart, and set the selectors to perform hard selections. We then measured EstiNet accuracy on the test set. EstiNet demonstrated significant performance improvement in inference mode compared to test mode. This improvement reveals the main potential advantage of EstiNet over existing approaches.

\section{EXPERIMENT DETAILS}
We applied the Estimate and Replace approach to a generated table-based, question-answering task (TAQ). We ran a supporting experiment on an auxiliary dataset to demonstrate the ability of Estimate and Replace to learn from less data. In the following section, we provide the TAQ task description and a detailed implementation of the EstiNet model that solves it.

\subsection{The TAQ Task}
For the TAQ task we generated a table-based question answering dataset. The TAQ dataset input has two parts: a question and a table. To correctly answer a question from this dataset, the DNN has to access the right table column and apply non-differentiable logic on it using a parameter it extracts from the query. For example, consider a table that describes the number of medals won by each country during the last Olympics, and a query such as: "Which countries won more than 7 gold medals?" To answer this query the DNN has to extract the argument (7 in this case) from the query, access the relevant column (namely, gold medals), and execute the ’greater than’ operation with the extracted argument and column content (namely a vector of numbers) as its parameters. The operation's output vector holds the indexes of the rows that satisfy the logic condition (greater-than in our example). The final answer contains the names of the countries (i.e., from the countries column) in the selected rows.

\subsubsection{TAQ API} 
Solving the TAQ task requires five basic logic functions: equal-to, less-than, greater-than, max, and min. Each such function defines an API that is composed of two inputs and one output. The first input is a vector of numbers, namely, a column in the table. The second is a scalar, namely, an argument from the question or NaN if the scalar parameter is not relevant. The output is one binary vector, the same size as the input vector. The output vector indicates the selected rows for a specific query and thus provides the answer.

\subsubsection{TAQ Data}
We generated tables in which the first row contains column names and the first column contains a list of entities (e.g., countries, teams, products, etc.). Subsequent columns contained the quantitative properties of an entity (e.g., population, number of wins, prices, discounts, etc.). Each TAQ-generated table consisted of 5 columns and 25 rows. 
We generated entity names (i.e., nations and clubs) for the first column by randomly selecting from a closed list. We generated values for the rest of the columns by sampling from a uniform distribution. We sampled values between 1 and 100 for the train set tables, and between 300 and 400 for the test set tables.
We further created 2 sets of randomly generated questions that use the 5 API functions. The set includes 20,000 train questions on the train tables and 4,000 test questions on the test tables. 

\subsection{Implementation Details}
In this section we provide more details on the exact implementation of the various EstiNet model parts that we designed to solve the TAQ task.
\subsubsection{Input Representation}
The TAQ input was composed of words, numbers, queries, and tables. The following describes the EstiNet representation for each of these elements.

\textbf{Word Representation:} EstiNet uses word pieces as in \cite{wu2016google} to represent words. A word is a concatenation of word pieces. EstiNet represents each word $w_j \in \mathbb{R}^d$ as an average value of its piece embedding.  

\textbf{Number Representation:} EstiNet aims to accurately represent a number and embed it into the same word vector space. Thus, its number representation follows the float32 scheme \cite{kahan1996ieee}. Specifically, it starts by representing a number $a\in \mathbb{R}$ as a 32 dimension Boolean vector $s'_n$. It then adds redundancy factor $r, r*{32}<d$ by multiplying each of the $s'_n$ digits $r$ times. Last, it pads the $s_n\in \mathbb{R}^d$ resulting vector with $d-r*{32}$ zeros.
We tried several representation schemes. This approach resulted in the best EstiNet performance.

\textbf{Query Representation:} 
EstiNet represents a query $q$ as a matrix of word embeddings and uses the LSTM model \cite{hochreiter1997long} to encode the query matrix into a vector representation: $q_{lstm} \in \mathbb{R}^{d\_rnn} = h_{last}(LSTM(Q))$ where $h_{last}$ is the last LSTM output and $d\_rnn$ is the dimension of the LSTM.

\textbf{Table Representation:} For the TAQ task, EstiNet represents a table $T \in \mathbb{R}^{n \times m \times d}$ with $n$ rows and $m$ columns as a three dimensional tensor. It represents a cell in a table as the piece average of its words.   

\subsubsection{Selectors} 
The EstiNet TAQ model uses three selector types: operation, argument, and column. Operation selectors select the correct API. Argument selectors select an argument from the query and hand it to the API. The column selector's role is to select a column from the table and hand it to the API.
EstiNet implements each selector subnetwork as a classifier. Let $C \in \mathbb{R}^{c_n \times d_c}$ be the predicted class matrix, where the total number of classes is $c_n$ and each class is represented by a vector of size $d_c$. For example, for a selector that has to select a word from a sentence, the $C$ matrix contains the word embeddings of the words in the sentence.
One may consider various selector implementation options. We use a simple, fully connected network implementation in which $W \in \mathbb{R}^{d\_rnn\ \times c_n}$ is the parameter matrix and $b \in \mathbb{R}^{d_c} $ is the bias. 
We define $\beta = C \times \left(q_{lstm} \times W + b\right)$ to be the selector prediction before activation and $\alpha = f_{sel}(.) = softmax(\beta)$ to be the prediction after the softmax activation layer. At inference time, the selector transforms its soft selection into a hard selection to satisfy the API requirements. EstiNet enables that using Gumbel Softmax hard selection functionality.
\subsubsection{EstiLayers}
The EstiNet TAQ model uses five EstiLayers to estimate each of the five logic operations. Each EstiLayer is a general purpose subnetwork that we implement with a transformer network encoder \cite{vaswani2017attention}. Specifically, we use \(n \in \mathbb{N}\) identical layers, each of which consists of two sub-layers. The first is a multi-head attention with \(k \in \mathbb{N}\) heads, and the second is a fully connected feed forward two-layer network, activated separately on each cell in the sequence. We then employ a residual connection around each of these two sub-layers, followed by layer normalization. Last, we apply linear transformation on the encoder output, adding bias and applying the Gumbel Softmax.

\section{RESULTS}
In this section we report on the performance of Estimate and Replace interacting with external applications. We start with the performance of the EstiNet model on the TAQ task. We then provide the performance results of offline and online training procedures and compare them with the plain training baseline. Last, we demonstrate the advantages of Estimate and Replace on learning from less data. 
\subsection{TAQ Performance}
\begin{figure*}
\centering     
\subfigure{\label{TAQ-task-results}\includegraphics[width=43mm]{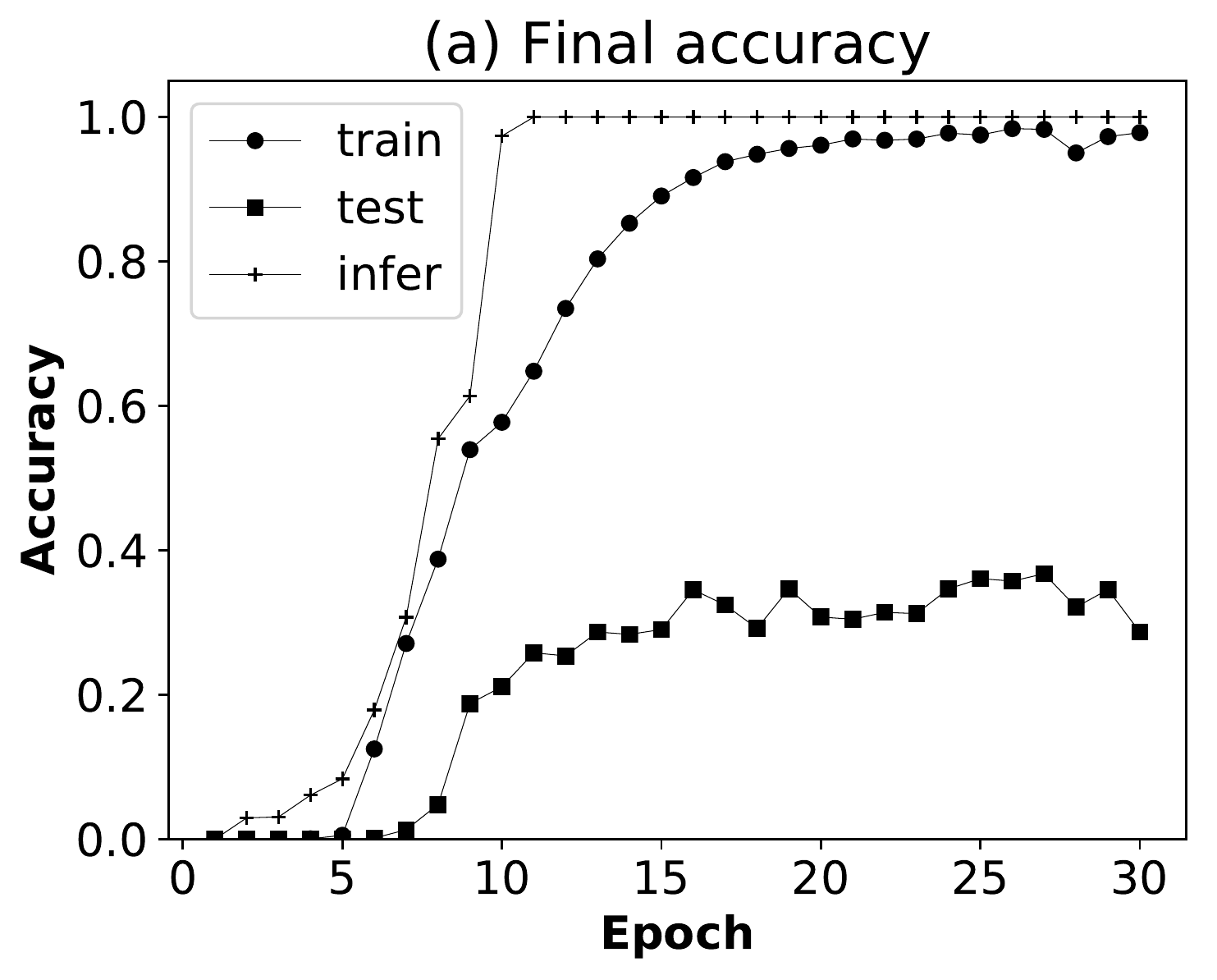}}
\subfigure{\label{fig:sel_train}\includegraphics[width=43mm]{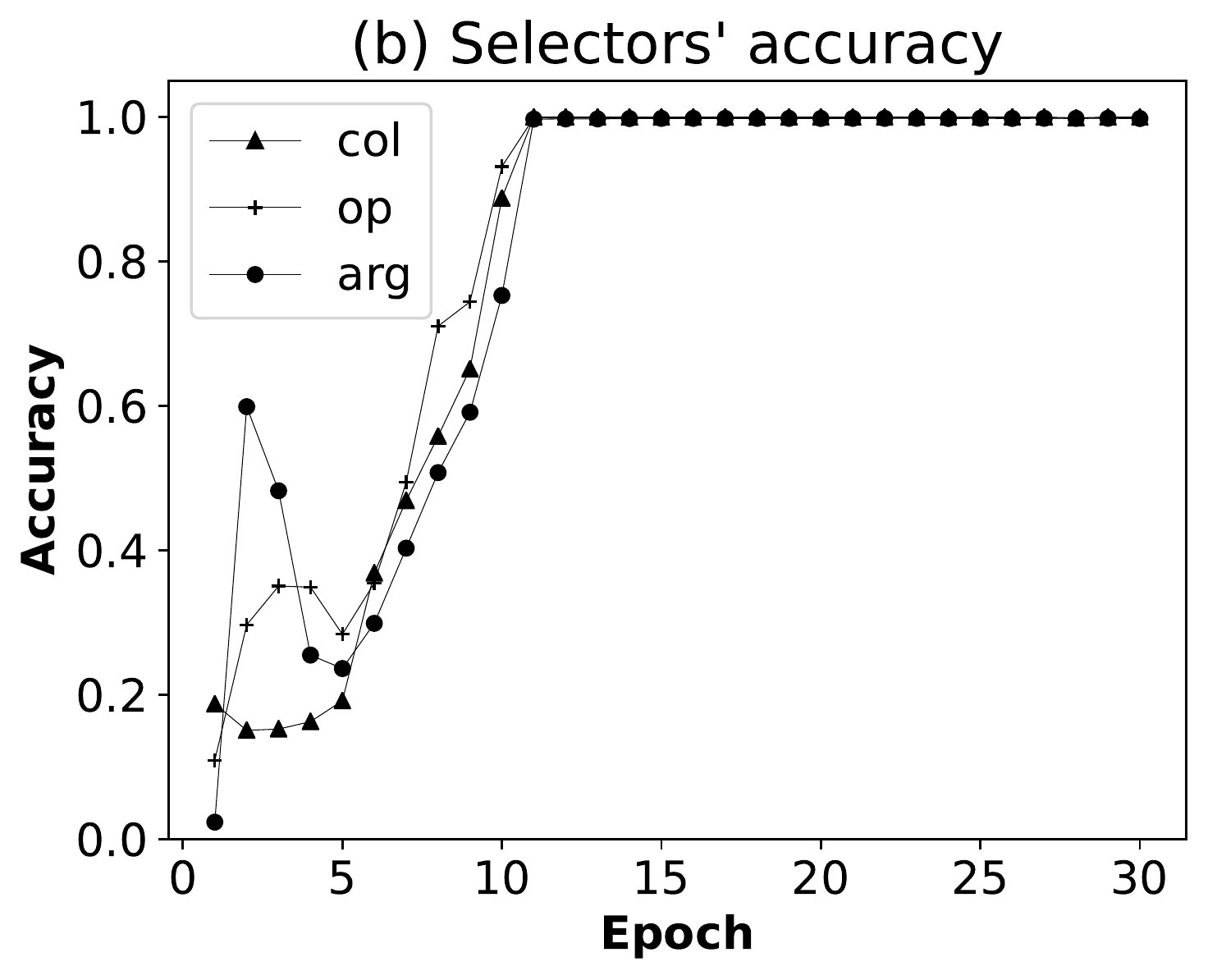}}
\subfigure{\label{fig:logic_train}\includegraphics[width=43mm]{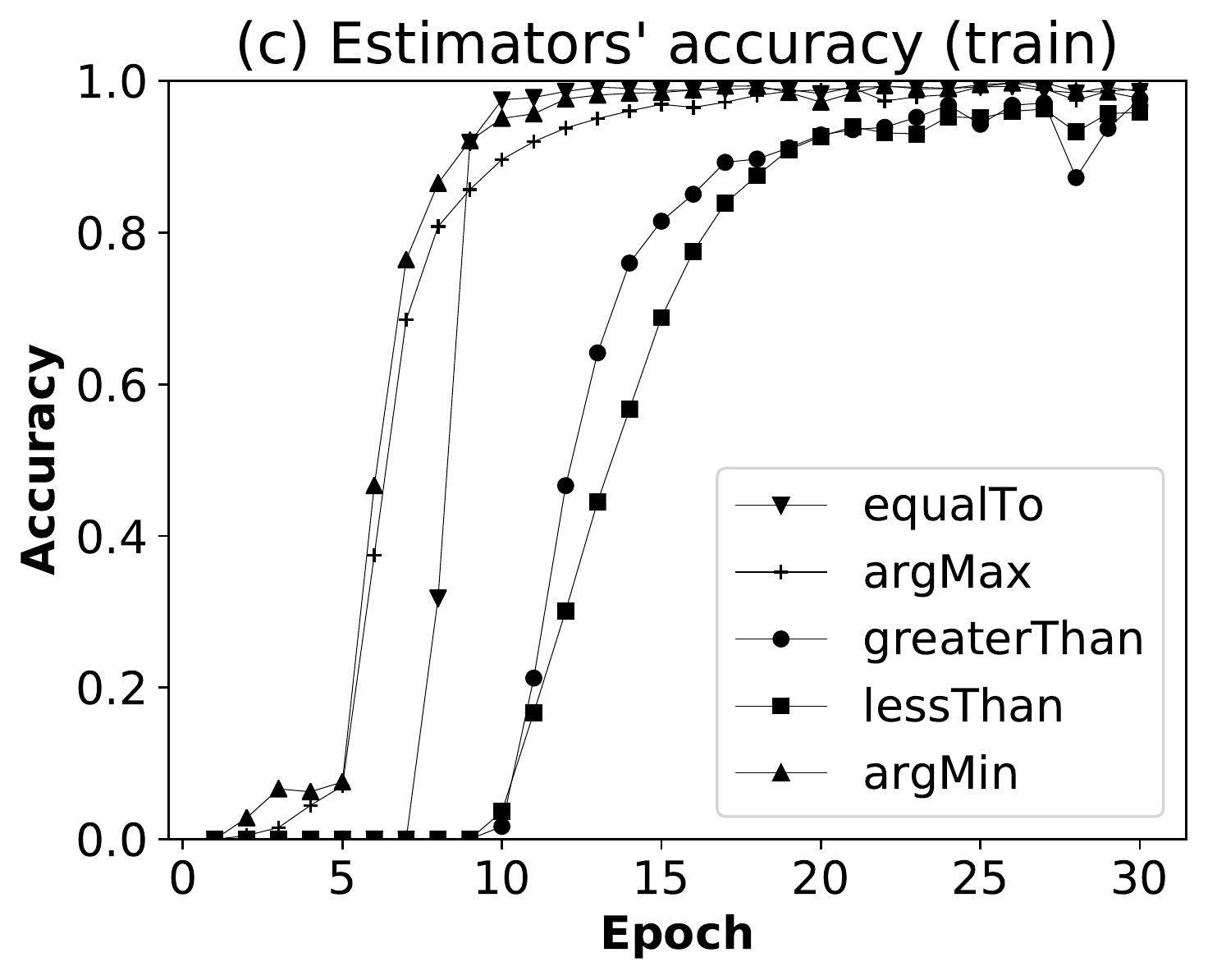}}
\subfigure{\label{fig:logic_test}\includegraphics[width=43mm]{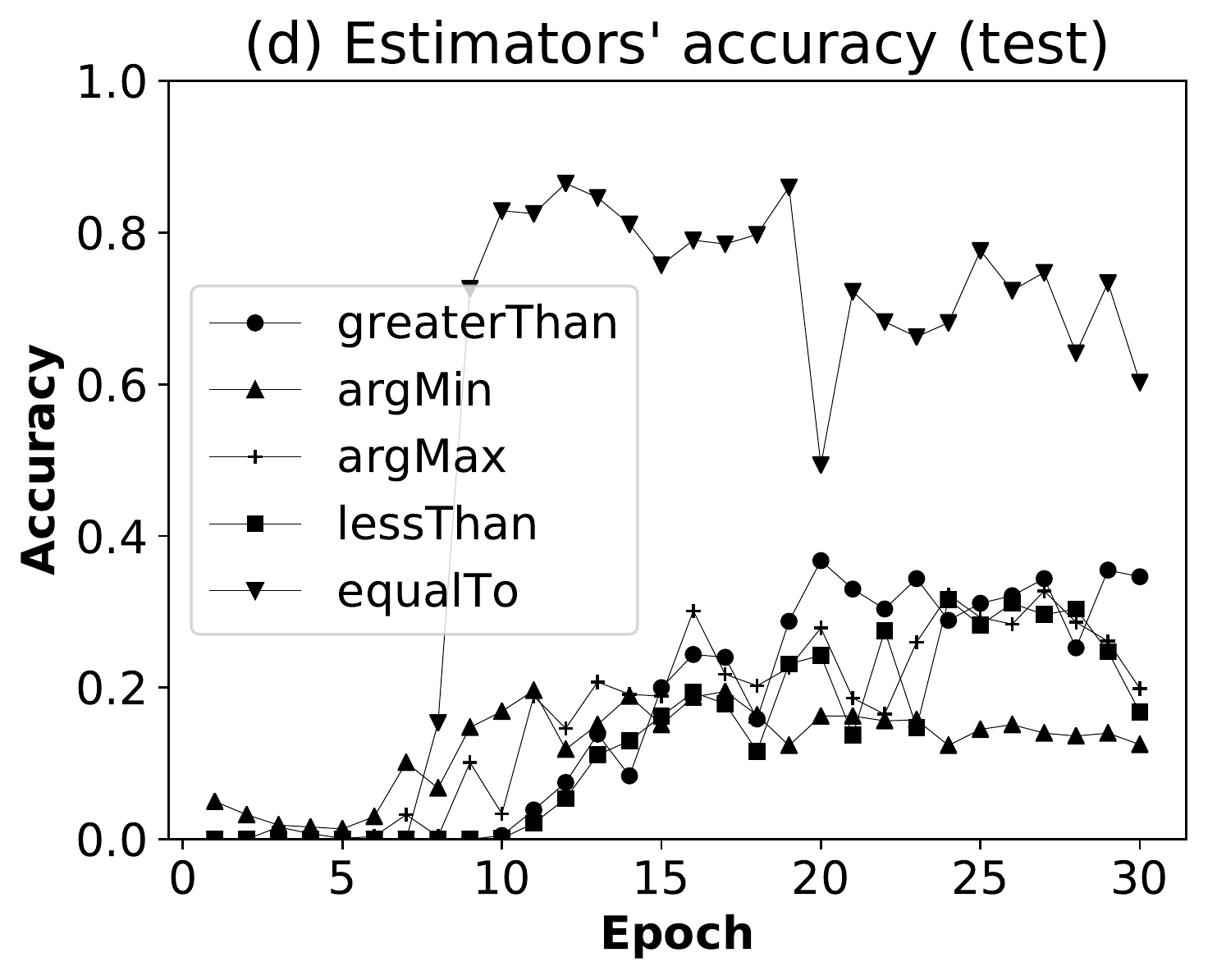}}
\caption{TAQ performance results - (a) Final accuracy in train, test, and inference modes. (b) Selector accuracy in train mode. (c) Estimator accuracy in train mode. (d) Estimator accuracy in test mode.}\label{fig:TAC_res}
\end{figure*}

Figure~\ref{TAQ-task-results} depicts the TAQ accuracy of the three model configurations: 1. Train: train model and train dataset 2. Test: train model and test dataset and 3. Inference: inference model and test dataset. As shown, the train model accuracy on the train dataset reaches 95\% after 20 epochs. Interestingly, inference accuracy on the test set is even higher and reaches 100\%, while test accuracy is lower than 40\%. The graph demonstrates the ability of the TAQ model to learn the logic interface despite its low generalization, indicated by the low test accuracy. Intermediate selector labels, which only exist for generated data, allow us to further assess selector accuracy in learning the interface with the estimators. Figure~\ref{fig:sel_train} presents the per-epoch accuracy of the three selectors during model training. The figure shows the selectors’ ability to perfectly learn the EstiLayers' interface after approximately 10 epochs. Selector test performance indicates the same (not shown in the figure). Figures~\ref{fig:logic_train}-~\ref{fig:logic_test} further assess EstiLayer accuracy in estimating the logic operation. The figures show that EstiLayers achieve near perfect accuracy on the train set, while test set performance is way below optimal. Most importantly, even though suboptimal EstiLayer test performance drives the overall low model performance on the test set, it has no effect on model performance while in inference mode. Clearly, this is because during inference mode we replace each EstiLayer with its application counterpart.

\subsection{Training Procedures}
\label{training_procedure_results}
Table~\ref{table:train_modes} summarizes the TAQ model performance throughout the training procedures described in Section \ref{training}. In plain training, which we use as our baseline (first line in the table), we train the TAQ model end-to-end without any additional constraints on the selector-estimator interface. As shown, the model can overfit the train set (0.9) but test set performance is low (0.11). Note that inference performance has no real meaning in that case, as there is no predefined selector-estimator interface. Thus, replacing the estimator with the external API makes no sense. Our main goal is to force a model to learn a predefined interface. To achieve that, we envision a two-step training procedure, which we refer to as offline training. We first train the estimators to overfit the train set and then use the trained estimator model during the end-to-end training process. The second line in the table shows the result of the offline experiment. As shown, the model fails to fit the train set and shows only 0.09 accuracy. Moreover, low train model accuracy results in low inference performance (0.17). We hypothesized that fixing the estimator parameters during the end-to-end training process prevents the rest of the model from fitting the train set. Thus, we ran a third experiment, offline-trainable, which allows the optimizer to update EstiLayer parameters during the end-to-end training process. This enabled the model to successfully fit the train set (0.97) but derogated the interface accuracy at inference time (0.33). With that failure in mind, we looked for a way to let the model learn the TAQ task end-to-end and at the same time force the estimator to comply with a predefined interface. We designed the online training procedure as a multi-task optimization process with two unrelated loss functions: 1) the task and 2) the external functionality (see Section \ref{online_training} for more details). In our first online training experiment (online entry in the table), we trained the estimators and the entire model without first initializing the estimator parameters. We assumed the estimators would gain their estimation functionality during the multi-task optimizing process. Indeed, this training procedure led to significant improvement in inference performance (0.69). However, the end-to-end model still faced difficulties in fitting the train set, gaining only 0.76 accuracy. To improve the end-to-end model learning performance, we experimented with the online-pretraining procedure (fifth line in the table), in which we started the end-to-end training process with trained estimator models. As indicated by the table, the improved online training procedure succeeded in fitting the train set (0.98) and achieved near perfect results at inference time (0.98). The low test set performance (0.47) is a phenomenon that usually indicates suboptimal model performance, (i.e., due to lack of data). However, in our case it demonstrates high generalization abilities in situations that lack data.

\begin{table} 
\small
\centering
\begin{tabular}{| l | c | c | c |}
\hline \bf Training & \bf Train & \bf Test & \bf Infer
\\  
\bf mode & \bf accuracy & \bf accuracy & \bf accuracy
\\  \hline
\bf Plain & 0.9 & 0.11 & 0.08\\
\bf Offline & 0.09 & 0.02 &  0.17\\
\bf Offline-trainable & 0.97 & 0.22 &  0.33\\
\bf Online & 0.76 & 0.22 &  0.69\\
\bf Online-pretraining & 0.98 & 0.47 &  0.98\\
\hline
\end{tabular}
{\caption{{\label{Training_modes_comparison_results} The table presents the accuracy of the three model configurations: train, test, and inference with the three training procedures: plain, offline, and online. Offline-trainable is offline mode in which we allow the optimizer to continue to update EstiLayer parameters. Online-pretraining is online mode in which we pre-trained the EstiLayers as in the offline mode. Each value in the table is calculated as an average of 10 repeated experiments.}}\label{table:train_modes}}
\end{table}

\subsection{Learning from Less Data}
To demonstrate the ability of Estimate and Replace to learn from less data, we ran an auxiliary experiment on a simpler dataset of greater-than/less-than questions. The questions came from 10 different templates, all requiring a true/false answer for 2 real numbers. For example: \textit{Out of x and y, is the first bigger ?} where $x,y$ are float numbers sampled from a $\sim \mathcal{N}(0,\,10^{10})$ distribution. 
The aim of this simple dataset was to demonstrate the ability of EstiNet to learn from less data. We compared the performance of the EstiNet model in plain and online training procedures. The plain training procedure served as a baseline and we measured its performance at test mode. This is because with plain training, the model does not learn to interact with  external applications, thus, inference performance has no meaning. On the other hand, online training lets the DNN model learn to interact with external applications, thus, we can measure its inference performance. Our experiment contained 5 train sets with 250, 500, 1,000, 5,000 or 10,000 questions and a test set with 1,000 questions. Table \ref{less-data-results-table} summarizes the performance differences. The results show that with online training the model generalizes better and accuracy differences between the two training procedures increase as the amount of training data decreases. It is interesting to note that to achieve, for example, 0.97 accuracy, the online training only needs samples that are 5\% of the data the plain training needs. We attribute the superiority of the EstiNet online training performance to its learning abilities. The model learns to interact with an external application to solve the logical part of the question.

\begin{table} 
\small
\centering
\begin{tabular}{| l | c | c | c | c | c |}
\hline \bf Train set size & \bf 250 & \bf 500 & \bf 1,000 & \bf 5,000 & \bf 10,000\\
\hline\bf Plain  & 0.533 & 0.686 & 0.859 & 0.931 &  0.98\\
\hline\bf Online  & 0.966 & 0.974 & 0.968 & 0.99.5 &  1.0\\
\hline\bf Difference & 81\% & 41\% & 13\% & 7\% &  2\%\\
\hline
\end{tabular}
\caption{The LWP task - The accuracy on test data using plain training and online training on different amounts of train data.}
\label{less-data-results-table}
\end{table}



\section{CONCLUSIONS AND FUTURE WORK}
Our work presents a new approach to overcoming the non-differentiability challenge, while integrating existing applications with DNNs. We use EstiLayers to estimate the external non-differentiable functionality. We then train the DNN end-to-end with the EstiLayers as functionality place -holders. The DNN learns the interface with the EstiLayers and uses it at inference time. Defining and learning an application interface, as well as learning when to use it, involves several non-differentiable aspects that must be overcome by the training process. These include hard selection, typed arguments, and functionality orchestration. We successfully demonstrated the advantages of EstiNet in learning and using a set of predefined, non-differentiable interfaces. We plan to follow-up on this work and extend it in two related directions. One direction is to dynamically learn the interface signature of a given functionality. Another is to learn to orchestrate a set of interfaces to solve a complex task. Achieving these will allow us to apply Estimate and Replace to real-world problems such as challenges in the finance and elementary science domains.
\bibliographystyle{named}
\bibliography{ijcai18}

\end{document}